\documentclass[conference]{IEEEtran}
\IEEEoverridecommandlockouts

\usepackage{amsmath,amssymb}
\usepackage{algorithmic}
\usepackage{textcomp}
\usepackage[dvipsnames]{xcolor}
\usepackage[normalem]{ulem} 
\def\BibTeX{{\rm B\kern-.05em{\sc i\kern-.025em b}\kern-.08em
    T\kern-.1667em\lower.7ex\hbox{E}\kern-.125emX}}
\usepackage{dashrule}
\usepackage{xurl}

\usepackage{tikz}         
\usepackage{pgfplots}     
\pgfplotsset{compat=1.18}
\usepgfplotslibrary{groupplots}
\usepackage{graphicx}     

\title{
    Real-Time Inference for Distributed Multimodal Systems under Communication Delay Uncertainty
    \thanks{
    This work was supported 
    by the Villum Investigator Grant ``WATER'' from the Velux Foundation, Denmark, and by
    the SNS JU project 6G-GOALS under the EU's Horizon Europe program under Grant Agreement No~101139232.}
}

\author{
    \IEEEauthorblockN{Victor Croisfelt$^{\dagger}$, João Henrique Inacio de Souza$^{\dagger}$, Shashi Raj Pandey$^{\dagger}$, Beatriz Soret$^{\ddagger\dagger}$, Petar Popovski$^{\dagger}$}
    \IEEEauthorblockA{{}$^{\dagger}$\textit{Department of Electronic Systems}, \textit{Aalborg University}, Denmark. E-mail: \{vcr,jhids,srp,petarp\}@es.aau.dk\\
    \textit{{}$^{\ddagger}$Telecommunications Research Institute}, \textit{Universidad de Málaga}, Spain. E-mail: bsa@uma.es}
}

\definecolor{amaranth}{rgb}{0.9, 0.17, 0.31}

\definecolor{coloraudio}{RGB}{0,127,255}  
\definecolor{colorvideo}{RGB}{255,181,112} 

\usepackage[acronym,,nohypertypes={acronym,notation}]{glossaries}
\newacronym{ma}{MA}{multimodal aggregator}

\newacronym{6g}{6G}{sixth generation}

\newacronym{fifo}{FIFO}{first-in first-out}

\newacronym{ai}{AI}{artificial intelligence}

\newacronym{cu}{CU}{control unit}

\newacronym{ap}{AP}{access point}
\newacronym{bs}{BS}{base station}

\newacronym{iot}{IoT}{internet of things}
\newacronym[longplural={temporal windows of integration}, shortplural={TWIs}]{twi}{TWI}{temporal window of integration}
\newacronym{dctw}{DCTW}{deep canonical time warping}

\newacronym{ar}{AR}{auto-regressive}

\newacronym{cpu}{CPU}{central processing unit}

\newacronym{ab}{AB}{auditory buffer}
\newacronym{vb}{VB}{visual buffer}

\newacronym{ml}{ML}{machine learning}

\newacronym{av}{AV}{audio-visual}
\newacronym{ave}{AVE}{AV event}
\newacronym{avel}{AVEL}{AV event localization}

\newacronym{lstm}{LSTM}{Long Short-Term Memory}

\newacronym{pmf}{PMF}{probability mass function}

\newacronym{atpf}{ATPF}{adaptive perception temporal framework}

\newacronym{ae}{AE}{auditory encoder}
\newacronym{ve}{VE}{visual encoder}

\newacronym{eta}{ETA}{Explicit Time Alignment}
\newacronym{ita}{ITA}{Implicit Time Alignment}

\newacronym{dsp}{DSP}{digital signal processing}
\newacronym{fpu}{FPU}{float point unit}

\newacronym{gan}{GAN}{Generative Adversarial Network}

\newacronym{mec}{MEC}{Mobile-Edge Computing}

\newacronym{mllm}{MLLM}{Multimodal Large Language Model}
\newacronym{mim}{MMLM}{Multimodal Machine Learning Model}

\newacronym{pts}{PTS}{Presentation Timestamp}

\newacronym{ack}{ACK}{acknowledgment}
\newacronym{snr}{SNR}{signal-to-noise ratio}
\newacronym{sota}{SotA}{state-of-the-art}

\newacronym{stft}{STFT}{short-time Fourier transform}

\newacronym{cnn}{CNN}{convolutional neural network}
\newacronym{rnn}{RNN}{recurrent neural network}

\newacronym{ntp}{NTP}{Network Time Protocol}
\newacronym{ptp}{PTP}{Precision Time Protocol}
\newacronym{gps}{GPS}{Global Positioning System}
\newacronym{rtp}{RTP}{Real-Time Transport Protocol}
\newacronym{rtcp}{RTCP}{RTP Control Protocol}
\newacronym{udp}{UDP}{User Datagram Protocol}

\newacronym{psv}{PSV}{probability of simultaneity violation}

\newacronym{embb}{eMBB}{enhanced mobile broadband}
\newacronym{emtc}{eMTC}{enhanced machine-type communication}
\newacronym{nbiot}{NB-IoT}{narrowband internet-of-things}

\begin{document}

\maketitle

\begin{abstract}
    Connected cyber-physical systems perform inference based on real-time inputs from multiple data streams. Uncertain communication delays across data streams challenge the temporal flow of the inference process. State-of-the-art (SotA) non-blocking inference methods rely on a \textit{reference-modality paradigm}, requiring one modality input to be fully received before processing, while depending on costly offline profiling. We propose a novel, \textit{neuro-inspired non-blocking inference paradigm} that primarily employs adaptive temporal windows of integration (TWIs) to dynamically adjust to stochastic delay patterns across heterogeneous streams while relaxing the reference-modality requirement. Our communication-delay-aware framework achieves robust real-time inference with finer-grained control over the accuracy--latency tradeoff. Experiments on the audio-visual event localization (AVEL) task demonstrate superior adaptability to network dynamics compared to SotA approaches.
\end{abstract}

\glsresetall 

\section{Introduction}
\label{sec:intro}
\IEEEPARstart{A}{s} \gls{ai} agents become increasingly ubiquitous across technological ecosystems, cyber-physical systems are emerging that tightly integrate physical processes with digital computation and networking. These distributed multimodal systems perform real-time inference by processing multimodal sensor inputs, such as visual, auditory, and LiDAR~\cite{lecun2024path,Popovski2024}. This deep integration supports autonomous decision-making in dynamic environments and spans various applications such as autonomous vehicles, industrial robotics, healthcare monitoring, and digital twins. Leveraging \glspl{mim} within these systems enhances their ability to deliver more accurate and resilient predictions by exploiting the complementary information inherent in multiple data modalities, beyond what unimodal models can offer~\cite{multimodal_survey}.

The human brain is known to dynamically perceive and integrate multimodal information across varying timescales~\cite{Vroomen_Keetels_2010}, achieving \textit{temporal coherence}, that is, the alignment and preservation of a consistent time-based relationship among data streams from different modalities. In other words, temporal coherence ensures that signals from multiple sensors remain synchronized and temporally correlated, so events occurring simultaneously (or with known delays) are perceived as unified in time. Similar to the human brain, distributed multimodal systems face the challenge of maintaining a coherent perception of time across modalities. Sensor streams are often acquired and transmitted by heterogeneous sources to a \gls{ma}, \emph{e.g.}, an edge \gls{ai} agent residing at an access point~\cite{Popovski2024}, while being subject to modality-specific, stochastic communication delays. These uncertain delays are often highly asymmetric, arising from fluctuating network conditions, bandwidth variability, packet loss, and the distinct characteristics of both the modalities and their underlying infrastructures~\cite{Popovski2024}. Maintaining temporal coherence ensures the system preserves a reliable sense of ``when'' events occur relative to each other, despite these uncertainties. This complexity raises a fundamental engineering challenge: \emph{How can real-time inference effectively handle asymmetric, uncertain communication delays in multimodal data streams while preserving robust temporal coherence?}

\begin{figure}[t]
    \noindent
    \includegraphics[scale=.99]{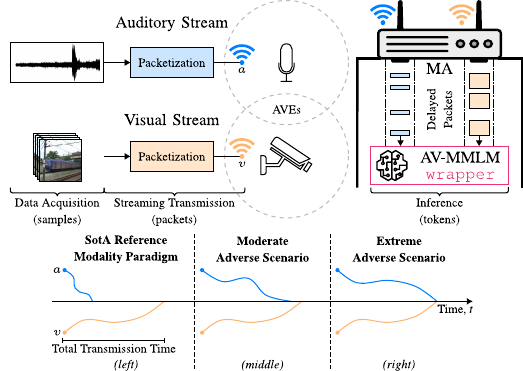}
    \vspace{-2ex}
    \caption{
        A distributed \gls{av} system streams unimodal auditory and visual data covering overlapping \glspl{ave}. At the \gls{ma}, a \texttt{wrapper} enables non-blocking inference by aligning delayed multimodal packets with a pre-trained token-based \gls{av}-\gls{mim} pipeline. Limitations of reference-modality SotA methods~\cite{Li2021SpeculativeInference,Wang2023PATCH,Wu2024AdaFlow,Xu2024MLLMInference} are demonstrated via two adverse scenarios.
    }
    \label{fig:system-model}
\end{figure}

{Existing approaches to this problem predominantly employ \emph{non-blocking inference} strategies at the \gls{ma}, operating in a plug-in manner by \texttt{wrapping} an \gls{mim} to incrementally update predictions as partial multimodal data arrive, while enforcing temporal coherence. The \gls{sota} \texttt{wrappers}}~\cite{Li2021SpeculativeInference,Wang2023PATCH,Wu2024AdaFlow,Xu2024MLLMInference} typically follow a \textit{reference-modality paradigm}, where inference is triggered only after fully receiving the input from a designated reference modality, usually the earliest arriving stream. Temporal coherence is ensured through a sequence of \gls{ml} modules conditioned on this reference, with parameters updated online using banks trained on human-curated, model-free offline profiles that capture expected source coding and communication characteristics that the system may face. Although this paradigm guarantees predictable performance and data integrity, it introduces practical limitations: it presumes constant and reliable availability of the reference modality, an assumption often violated in dynamic or lossy networks; its rigid temporal dependence restricts the exploitation of partial observations, limiting early or subtle decision-making; and changing the reference modality to adapt to degradation requires extensive retraining and additional memory overhead, ultimately compromising flexibility and scalability.

To illustrate these limitations concretely, Fig.~\ref{fig:system-model} depicts a canonical bimodal \gls{av} distributed system, where unimodal auditory and visual streams are captured by two separate sources, each independently streaming continuous modality data to the \gls{ma}. These transmissions are subject to modality-specific communication delay uncertainties. At the \gls{ma}, a pre-trained \gls{av}-\gls{mim} supports tasks such as \gls{avel}~\cite{Tian2018AVE,Zhou2025Towards}, where an \gls{ave} corresponds to an event that is both audible and visible within a scene. Consistent with \gls{sota} practices (Fig.~\ref{fig:system-model}, \textit{left}), the auditory stream is typically treated as the reference modality due to its comparatively lightweight nature relative to the bandwidth-heavy visual stream. These methods often assume near-instantaneous audio reception. However, despite its low bandwidth consumption, communication uncertainties can significantly influence how the \gls{ma} perceives and synchronizes multimodal data. In a \underline{moderate adverse scenario} with respect to communication delay uncertainty (Fig.~\ref{fig:system-model}, \textit{middle}), the transmission time of the reference stream approaches that of the non-reference modality, leading to substantial inference delays because the system stalls until full reference reception. As the time difference between modalities shrinks, the effectiveness of \gls{sota} approaches diminishes. In an \underline{extreme adverse scenario} (Fig.~\ref{fig:system-model}, \textit{right}), the communication delays for both modalities converge or become comparable, making reference selection ambiguous and fully compromising \gls{sota} performance. These challenges can often arise in practice from highly variable network conditions and heterogeneous source resources; practical factors often overlooked in prior work.




%
\subsection{{Contributions}}
In this paper, we propose a novel \textit{neuro-inspired non-blocking inference} paradigm aimed at overcoming the limitations of reference-modality approaches~\cite{Li2021SpeculativeInference,Wang2023PATCH,Wu2024AdaFlow,Xu2024MLLMInference}.  Inspired by neuroscientific findings on how the human brain maintains temporal coherence across asynchronous modalities~\cite{Vroomen_Keetels_2010,Popovski2024}, our method leverages the concept of \glspl{twi}, adaptive temporal windows that flexibly correct the binding of temporally misaligned data streams by, for example, expanding to accommodate delays when a modality lags another, enabling accurate multimodal integration. 

{We start by observing that existing \gls{sota} \texttt{wrappers}~\cite{Li2021SpeculativeInference,Wang2023PATCH,Wu2024AdaFlow,Xu2024MLLMInference} do not model communication as an integral part of the system; instead, communication is treated as an external or predetermined factor, overlooking latency, variability, and delays within the inference process.} Motivated by this, we introduce a system model that explicitly models the multimodal streaming data source as a composition of unimodal streams while statistically characterizing their communication delay uncertainties in alignment with application-specific requirements of the \gls{mim}. This model-driven framework provides predictive capabilities to statistically anticipate the arrival of modality-specific data. We further observe that streaming distributed multimodal inference under \gls{sota} \glspl{mim}~\cite{multimodal_survey}, data is processed at three distinctive granularities: sample (acquisition), packet (transmission), and token (inference). Packets encapsulate subsets of raw samples for transmission, whereas tokens aggregate fixed-length groups of samples into semantically meaningful units for inference. \textit{Accordingly, the essential role of a \texttt{wrapper} is to align packet streams from different modalities, each subject to distinct delay uncertainties, and generate temporally consistent token sequences.}

Building on these insights, we propose a neuro-inspired \texttt{wrapper} for real-time inference that maintains computational temporal coherence by synchronizing delayed unimodal streams through adaptive temporal windows of integration (\glspl{twi}) and auxiliary mechanisms within a tokenized \gls{mim} pipeline. These dynamic, statistically optimized TWIs, guided by predictive system modeling, enable flexible inference triggering based on data granularity and relax the requirement for complete reference-modality reception. Consequently, our approach enables operation along the accuracy–latency tradeoff curve, enhancing robustness to communication delays and improving resource utilization via parallel multimodal processing under varying network conditions.

\section{System Model}\label{sec:system-model}
Without loss of generality, we consider the bimodal \gls{av} inference setup illustrated in Fig.~\ref{fig:system-model}. We denote by $t\in\mathbb{R}_{+},\, t \to \infty$ the continuous \textit{physical time}, representing the unbounded, real-world timeline along which events occur in nature. This work focuses on the impact of communication delay variations on inference performance, while variations in acquisition and computation times are considered negligible.



\subsection{Multimodal Streaming Data Source}\label{sec:system-model:sources}
We model the observed process by discretizing \( t \) into \( N_o \) sequential video observations \( O_i \), which together form the multimodal streaming data source, where \( i \in \{1, 2, \ldots, N_o\} \) indexes the observations along time. Each observation spans a fixed temporal interval \( T_{\mathrm{video}} \)
, representing a localized snapshot of the underlying physical process. Every \( O_i \) consists of an auditory component \( O_{i,a} \), captured by, \emph{e.g.}, a remote microphone, and a visual component \( O_{i,v} \), acquired by, \emph{e.g.}, a surveillance camera; forming unimodal streaming data sources. We consider continuous, unbounded multimodal data streaming with \( N_o \to \infty \)~\cite{purushwalkam2022challenges}.


%
\noindent
\textbf{Per-Modality Packetization.}  
Each unimodal source acquires modality-specific samples and segments them into fixed-length packets for streaming transmission. For auditory and visual streams, the corresponding packets \( a_i \in \{0,1\}^{L_a} \) and \( v_i \in \{0,1\}^{L_v} \) have lengths \( L_a \) and \( L_v \) bits, respectively. Given generation rates \( R_a \) and \( R_v \) (samples per second) and source coding rates \( B_a \) and \( B_v \) (bits per sample), each packet spans durations \( D_a \!=\! \frac{L_a}{B_a R_a} \) and \( D_v\! =\! \frac{L_v}{B_v R_v} \) for the auditory and visual modalities, respectively. Consequently, an observation window of length \( T_{\mathrm{video}} \) comprises \( N_a \!=\! \lceil \frac{T_{\mathrm{video}}}{D_a} \rceil \) auditory packets and \( N_v\! =\! \lceil \frac{T_{\mathrm{video}}}{D_v} \rceil \) visual packets, where zero-padding is applied to fill any partial packets. The number of samples per packet is \( S_a^p \!=\! \frac{L_a}{B_a} \) and \( S_v^p \!=\! \frac{L_v}{B_v} \), allowing each packet to contain fractional samples. We assume perfect synchronization among sources and the \gls{ma}, achievable through external synchronization protocols such as the \gls{ntp}. Thus, each packet includes \glspl{pts} as metadata for temporal alignment, along with an observation index \( i \), whose sizes are negligible relative to the packet length \( L_s \) for \( s \in \{a,v\} \). Conventional source coding is employed as a baseline, without adopting semantic compression~\cite{shen2025compressionpixelssemanticcompression}, and packets are encoded independently to preserve maximum streaming flexibility.

\noindent
\textbf{Packet-View of the Multimodal Streaming Data Source.}
From a communication perspective, \( O_i \) is viewed as sequences of heterogeneous packets. Each observation \( O_i \!=\! (O_{i,a}, O_{i,v}) \) comprises modality-specific packet sequences, namely \( O_{i,a} \!= \! (a_i^{j_a}) \) for the auditory stream and \( O_{i,v} \!=\! (v_i^{j_v}) \) for the visual, indexed by \( j_a \in \{1, 2, \dots, N_a\} \) and \( j_v \in \{1, 2, \dots, N_v\} \), respectively. For the \( i \)-th observation, the end time of each auditory or visual packet, before transmission, can be timestamped onto the physical time \( t \) as:
\begin{equation}
    t_i^{j_a} = {(i-1)} T_{\rm video} + j_a D_a, 
    ~~~
    t_i^{j_v} = {(i-1)} T_{\rm video} + j_v D_v.
    \label{eq:packet-end-time}
\end{equation}
These packets represent discrete transmission units subject to modality-specific communication delay uncertainties, as characterized next.



\subsection{Communication Delay Uncertainties}\label{sec:system-model:communication-model}
We adopt the per-packet delay model from~\cite{Suman2023statistical}, where unimodal streams are transmitted over independent wireless channels to the \gls{ma} using time-slotted packet switching with one packet per slot. Each source is modeled as an infinite \gls{fifo} buffer with unlimited retransmissions and backlog. The sources and the \gls{ma} are equipped with a single antenna each. The \(j_s\)-th packet of the \(i\)-th observation in stream \(s\!\in\!\{a,v\}\) experiences Rayleigh block-fading per slot, with received \gls{snr} \(\gamma^{j_s}_i \sim \exp(1/\bar{\gamma}_s)\), where \(\bar{\gamma}_s\) denotes the average \gls{snr}. Under an erasure channel model with outage probability \(\varepsilon_s\), the effective transmission rate is \(\eta_s(\varepsilon_s) \!=\! W_s \log_2(1 - \bar{\gamma}_s \ln(1-\varepsilon_s))\), where $W_s$ is the bandwidth allocated to stream $s$. We consider a communication-constrained streaming regime where the source rate exceeds the communication rate, \(B_s R_s \gg \eta_s(\varepsilon_s)\), ensuring persistent buffer backlog. Assuming virtually perfect reliability through retransmissions and negligible per-packet \gls{ack} delay, the mean packet transmission duration is given by:
%
$
    \Gamma_s
    \!=\! \tfrac{L_s}{\eta_s(\varepsilon_s)}.
$
%
Accordingly, the stochastic communication delay \(T_{i}^{j_s}\) of each packet follows the \gls{pmf} with $R\in\mathbb{N},\, R\geq 1$ denoting {the number of transmissions until successful {packet} reception}:
\begin{equation}
    \Pr\{T_{i}^{j_s} = R \Gamma_s\} = \varepsilon_s^{R-1} (1-\varepsilon_s).
    \label{eq:communication-latency}
\end{equation}
Its mean and variance are $\mathbb{E}[T_i^{j_s}] \!=\! \tfrac{\Gamma_s}{1-\varepsilon_s}$ and $\mathrm{Var}[T_i^{j_s}] \!=\! \tfrac{\varepsilon_s(\Gamma_s)^2}{(1-\varepsilon_s)^2}.$ This delay model captures the communication uncertainty as a function of the parameters: SNR, bandwidth, and outage probability, which can vary across streams. 

\subsection{Abstracted Multimodal Machine Learning Model}\label{sec:system-model:inference-model}
\begin{figure*}[t]
    \centering
    \includegraphics[scale=1, trim=0cm 0cm 0cm 0pt, clip]{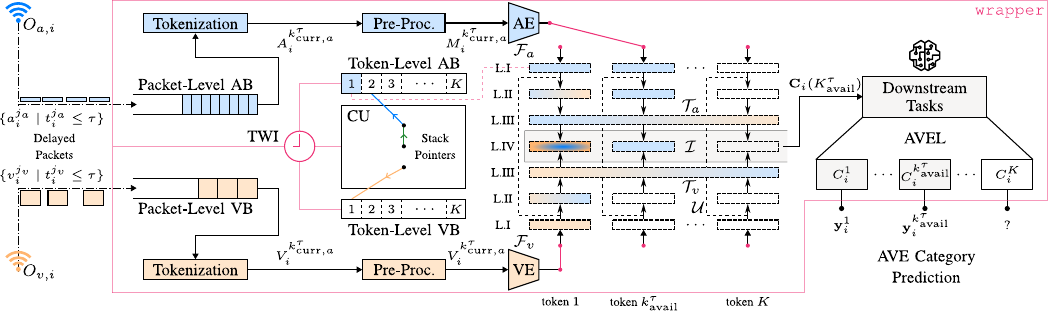}
    \vspace{-1em}
    \caption{
        Snapshot of a \texttt{wrapped} pre-trained \gls{av}-\gls{mim} at reception time \(\tau\). All \texttt{wrapper} operations are synchronized by a \gls{twi}-based clock. At \(\tau\), unimodal streaming data sources provide modality-specific packets—each containing a subset of input samples—to the \gls{ma}, affected by distinct communication delay uncertainties. The \texttt{wrapper} aligns asynchronous packets and converts them into token-based representations derived from the underlying samples for \gls{av}-\gls{mim} processing, employing mechanisms to ensure temporal coherence, including \gls{twi} optimization. In the figure, uncolored blocks represent missing data (zero-imputed), colored blocks indicate partial or complete data. Semantic embedding levels are denoted as L.I, L.II, L.III, and L.IV; AB and VB refer to auditory and visual buffers; AE and VE to auditory and visual L.I unimodal encoders; and CU to the control unit. For illustration, \(d_a = d_v = d_a' = d_v' = d\). 
    }
    \label{fig:inference-pipeline}
\end{figure*}

To define minimal temporal requirements for downstream applications, we focus specifically on a generic encoder-only model~\cite{Tian2018AVE,Mahmud2023AVECLIP,Zhou2025Towards} designed for the \gls{avel} task. Success in this task often relies on extended temporal context, motivating a hierarchical integration of multimodal information across two primary temporal scales: (i) the \emph{sample level}, which captures fine-grained spatiotemporal structure within pre-defined segments to produce tokens, and (ii) the \emph{token level}, where tokens function as atomic units for inference. At this token-level, \gls{av}-\gls{mim} extracts high-level semantic embeddings to accomplish the task. Figure~\ref{fig:inference-pipeline} illustrates a \texttt{wrapped} \gls{av}-\gls{mim}-based inference pipeline employing a generic four-tier hierarchy that derives high-level semantic embeddings through unimodal token encoding (\textbf{Level I}), early multimodal fusion (\textbf{Level II}), temporal token modeling (\textbf{Level III}), and late-stage multimodal fusion (\textbf{Level IV}).

\noindent
\textbf{Expected Multimodal Data Input and Tokenization.} 
The \gls{av}-\gls{mim} treats each observation \( O_i \) of fixed duration \( T_{\mathrm{video}} \) as an input, processing it holistically by internally partitioning it into \( K \) non-overlapping sequential tokens, each of duration \( T_k \), indexed by \( k \in \{1, 2, \dots, K\} \). We denote the \( k \)-th auditory and visual tokens as \( A_k \) and \( V_k \), respectively. Conventionally, each auditory token \( A_i^k \in [-1, +1]^{S_a \times 1} \) represents a 16-bit mono waveform, while each visual token \( V_i^k \in \mathbb{Z}^{S_v \times H \times W \times C} \) encodes 8-bit pixel frames of height \( H \), width \( W \), and number of color channels \( C \). The sample counts \( S_a = R_a T_k \) and \( S_v = R_v T_k \) are typically integer values, ensured by choosing \( T_k \) to match the generation rates \( R_a \) and \( R_v \). 

%
\noindent
\textbf{Token-View of the Multimodal Streaming Data Source.}  
Contrasting with the packet-level representation, the model views \(O_i\) as sequences of \gls{av} tokens. Each observation \(O_i \!= \!(O_{i,a}, O_{i,v})\) is represented internally as modality-specific token sequences, namely \(O_{i,a} \!=\! (A_i^k)\) for the auditory stream and \(O_{i,v} \!= \!(V_i^k)\) for the visual. To reconcile the packet and token views, it is noted that both packets and tokens are constructed from underlying samples, with tokens requiring a specific number of packets for completion, defined by
$N_a^k \!=\! \lceil \frac{T_k}{D_a} \rceil$ and $N_v^k \!=\! \lceil \frac{T_k}{D_v} \rceil.$ Within this token abstraction, token end times $t_i^k \!= \!{(i-1)} T_{\rm video} + k T_k$ establish a normalized temporal scale optimized for inference, which becomes unattainable due to communication delay uncertainties.



%
\noindent
\textbf{Inference Pipeline.}
The \gls{av}-\gls{mim} pipeline operates on synchronized multimodal observations \(O_i\), each segmented into \(K\) \gls{av} token pairs \((A_i^k, V_i^k)\). Auditory tokens \(A_i^k\) are first transformed into log mel-spectrograms \(M_i^k \in \mathbb{R}_+^{S_a' \times F_b}\), motivated by human-hearing capabilities~\cite{Tian2018AVE}. Visual tokens \(V_i^k\) represent RGB images compatible with \gls{cnn} backbones such as VGG or ResNet. After pre-processing, each modality is encoded through unimodal encoders, \(\mathcal{F}_a(M_i^k)\) and \(\mathcal{F}_v(V_i^k)\), producing \textbf{Level-I} feature embeddings \(F_{i,a}^k \in \mathbb{R}^{d_a}\) and \(F_{i,v}^k \in \mathbb{R}^{d_v}\), collected as sequences \(\mathbf{F}_{i,a} \!=\! (F_{i,a}^k)\) and \(\mathbf{F}_{i,v}\! =\! (F_{i,v}^k)\). Temporal modeling across tokens is achieved by modality-specific encoders \(\mathcal{T}_a\) and \(\mathcal{T}_v\), optionally preceded by early fusion mechanisms \(\mathcal{U}\) such as audio-guided visual attention~\cite{Tian2018AVE} or cross-modal alignment~\cite{Zhou2025Towards}, yielding temporally enriched embeddings \(E_{i,a}^k \!=\! \mathcal{T}_a(\mathcal{U}(\mathbf{F}_{i,a}))\) and \(E_{i,v}^k \!=\! \mathcal{T}_v(\mathcal{U}(\mathbf{F}_{i,v}))\) (\textbf{Level II and III}). Finally, late-stage multimodal fusion \(\mathcal{I}\) integrates both modalities to form \textbf{Level-IV} fused embeddings \(C_i^k \!=\! \mathcal{I}(E_{i,a}^k, E_{i,v}^k) \in \mathbb{R}^d\), aggregated into sequences \(\mathbf{C}_i\! =\! (C_i^k)\) that capture synchronized multimodal semantic cues for downstream tasks as \gls{avel}.

The presented system model represents the multimodal streaming data source as packets for transmission and as tokens for inference, both ultimately derived from samples acquired by unimodal sources. Crucially, the effectiveness of real-time \gls{av}-\gls{mim} inference hinges on precise synchronization at both the sample and token levels; failure to maintain this coherence would significantly degrade feature quality. This critical responsibility is managed by the \texttt{wrapper}. In the following section, we provide a formal problem formulation that captures this synchronization challenge.

\section{Problem Formulation}\label{sec:problem-formulation}
Drawing inspiration from self-supervised streaming learning~\cite{purushwalkam2022challenges}, we 
define a distributed multimodal inference problem while explicitly contrasting our objective with current \gls{sota} methods~\cite{Li2021SpeculativeInference,Wang2023PATCH,Wu2024AdaFlow,Xu2024MLLMInference}.
{For a given observation \( O_i \), let \( T_{i,\min} \) be the minimum end-to-end inference latency relating to the total transmission time of the fastest modality to reach the \gls{ma}}:  
\begin{equation}
    T_{i,\min} = \min_{s \in \{a,v\}} T_{i,s},
    \label{eq:minimum-end2end-latency}
\end{equation}
where \( T_{i,s} \!=\! \sum_{j_s=1}^{N_s} T_i^{j_s} \) denotes the total transmission time of all packets in stream \( s \) for observation \( i \), and each \( T_i^{j_s} \) follows~\eqref{eq:communication-latency}. We aim to design a \texttt{wrapper} that relaxes the reference-modality paradigm, enabling inference to start before complete reception of any modality, thus achieving latency below \( T_{i,\min} \).


Let $\tau$ denote the reception timeline at the \gls{ma}, where $\tau$ belongs to a discrete subset \(\mathcal{S} \subseteq \mathbb{R}_+\) of the physical time \(t \in \mathbb{R}_+\). 
At a given $\tau$, for observation \(O_i\), data fetched from unimodal sources \(O_{i,a}\) and \(O_{i,v}\) reveal subsets of samples contained in packets $
\{a_i^{j_a} \mid t_i^{j_a} \leq \tau\}$ and $\{v_i^{j_v} \mid t_i^{j_v} \leq \tau\},$
where, updating~\eqref{eq:packet-end-time}, the $j_{s}$-th \textit{packet reception time} at the \gls{ma} for stream \(s\in\{a,v\}\) can be timestamped as:
%
\begin{equation}
    t_i^{j_s} = (i-1) T_{\rm video} + D_s + {\sum_{j'={1}}^{{j_s}} T_i^{j^{\prime}}},
    \label{eq:packet-reception-time}
\end{equation}
where \(T_i^{j^{\prime}}\) follows~\eqref{eq:communication-latency}. Samples fetched at time \(\tau\) remain available if buffered, while unrevealed future samples are inaccessible. Due to unbounded streaming data, efficient buffer management is crucial. Variations in data and communication delays cause asynchronous arrivals across modalities; thus, samples corresponding to the same physical time \(t\) may arrive at different reception times \(\tau\), requiring mechanisms to ensure temporal coherence.

\section{Wrapper with Temporal Integration}\label{sec:wrapper}
To address the described problem, we propose a neuro-inspired \texttt{wrapper} that primarily employs \glspl{twi} as its internal clock source, featuring a dynamically adjustable integration period \( T_W \ll T_{i,\min} \). This mechanism enables time coherence for real-time inference in conjunction with auxiliary modules such as buffer management and control. The reception timeline progresses in discrete steps of \( T_W \), that is, \(\tau \in \mathcal{S}\) where \(\mathcal{S} = \{0, T_W, 2T_W, \dots\}\). Within each interval, the \texttt{wrapper} aggregates and buffers delayed multimodal packets from unimodal sources and generates updated predictions at the end of the period. In the following, we present an operational overview of the \texttt{wrapper}, illustrated in Fig.~\ref{fig:inference-pipeline}, followed by the statistical optimization process of the \gls{twi}.



\subsection{Operational Overview}

%
\noindent
\textbf{Buffer Management.}
We propose two per-modality buffer types, namely \gls{ab} and \gls{vb}: a packet-level buffer fetching incoming packets, and a token-level buffer that stores Level-I features from previously fully received and processed tokens of a given observation $i$. The packet-level buffer size varies with delay uncertainty (assumed of infinity size), while the token-level buffer has a fixed size \(\mathcal{O}(K d_s)\) for \(s\!\in\!\{a,v\}\); being indexed by token positions \(k \in \{1, 2,\dots, K\}\). The number of fully received tokens stored at reception time \(\tau\) is \(K^{\tau}_{a, \mathrm{full}} \leq K\) and \(K^{\tau}_{v, \mathrm{full}} \leq K\) for the auditory and visual streams, respectively.

\noindent
\textbf{Control Unit.}
We design a \gls{cu} to bridge packet reception and token processing, maintaining temporal coherence. At $\tau$, it tracks the current observation \(i\) with stack pointers \(k^{\tau}_{a,\mathrm{curr}}\) and \(k^{\tau}_{v,\mathrm{curr}}\), indicating the latest fully received tokens in buffers. A third pointer \(k^{\tau}_{\mathrm{avail}} = \max(k^{\tau}_{a,\mathrm{curr}}, k^{\tau}_{v,\mathrm{curr}})\) tracks available tokens \(K^{\tau}_{\mathrm{avail}} \leq K\), enabling asynchronous, partial-modality fusion.\footnote{
    Conversely, using the minimum pointer value would enforce inference only once both modalities are available. We use the maximum in the experiments.
} 

\noindent
\textbf{Operational Description.}
At reception time \(\tau\), arriving auditory and visual packets are buffered by modality. The \gls{cu} identifies the current observation using packet metadata, prioritizing the earliest available one. For the active observation, packets are tokenized by concatenating samples based on \glspl{pts} and incrementally pre-processed, enabling efficient memory management via caching. The \texttt{wrapper} executes the \gls{av}-\gls{mim} pipeline by fetching completed Level-I token embeddings from token-level buffers and applying feature extraction functions \(\mathcal{F}_a\) and \(\mathcal{F}_v\) to current (potentially partial) tokens per modality. Future samples and tokens are zero-imputed. All accessible tokens indexed by \(k^{\tau}_{\mathrm{avail}} \in \{1,2,\dots,K^{\tau}_{\mathrm{avail}}\}\) are processed to compute fused embeddings \(\mathbf{C}_i(K^{\tau}_{\mathrm{avail}})\) via the pipeline \(\mathcal{U} \rightarrow \mathcal{T}_s \rightarrow \mathcal{I}\). At each \gls{twi} conclusion, the system outputs token-based predictions, stores completed tokens 
if any, 
and prunes packets associated with completed tokens from the packet-level buffers. The active observation ends when 
\(K^{\tau}_{a, \mathrm{full}} = K^{\tau}_{v, \mathrm{full}} = K\), resetting buffers and pointers.

\subsection{Statistically Optimizing the Temporal Integration Period}
At the end of each integration period \(T_W\), the per-token predictions are obtained based on the fused embeddings \(\mathbf{C}_i(K^{\tau}_{\mathrm{avail}})\). The objective is to optimize \(T_W\) to determine the appropriate moment for triggering inference while considering communication delay uncertainties. To illustrate this concept, we propose a \textit{mean-based design strategy}. Specifically, \(T_W\) is defined as the minimum interval required, on average, to accumulate a ``sufficient'' amount of data for meaningful model execution; formally expressed as:
\begin{equation}
    T_W = \max_{s \in \{a,v\}} \mathbb{E}\!\left[\sum_{j_s=1}^{P_s} T_i^{j_s}\right],
    \label{eq:temporal-integration-optimization-unconstrained}
\end{equation}
where the expectation is taken over the random observations $(O_i)$ and packet arrivals {($T_i^{j_s}$)}. 
Two design variants are considered. Setting \(P_a \!=\! P_v \!=\! 1\) yields the \textit{At-Least One-Packet Per-Modality \textbf{(PaMo)} \gls{twi}}, ensuring that, on average, at least one packet per modality is received before inference. Alternatively, selecting \(P_a\! =\! N_a^k\) and \(P_v\! =\! N_v^k\) defines the \textit{At-Least One-Token Per-Modality \textbf{(ToMo)} \gls{twi}}, which ensures that, on average, each modality contributes at least one token before inference. Problem~\eqref{eq:temporal-integration-optimization-unconstrained} can be solved in closed form by noting that the random variables \(\{T_i^{j_s}\}_{j_s=1}^{N_s}\) are independent; hence, the expectation of their sum equals the sum of their expectations 
with the optimization reducing to:
%
$
    \textstyle
    T_{W,\text{PaMo}}^\star = \max \{ 
        \textstyle\frac{\Gamma_a}{1-\varepsilon_a},\;
        \textstyle\frac{\Gamma_v}{1-\varepsilon_v}
    \}
$ 
and
$
    \textstyle
    T_{W,\text{ToMo}}^\star = \max \{ 
        \textstyle\frac{ N_a^k\Gamma_a}{1-\varepsilon_a},\;
        \textstyle\frac{ N_v^k\Gamma_v}{1-\varepsilon_v}
    \}.
$

%
%


\section{Experiments}
\label{sec:experiments}
We consider the \gls{avel} dataset~\cite{Tian2018AVE} in a fully supervised setting. The dataset comprises 4097 \(T_{\rm video}\!=\!10\,\mathrm{s}\) video observations, partitioned into training (80\%), validation (10\%), and test (10\%) subsets.\footnote{Codes available on: \url{https://github.com/victorkreutzfeldt/real-time-inference-distributed-multimodal-systems}.} Each observation is segmented into $K\!=\!10$ non-overlapping \(T_k\!=\!1\,\mathrm{s}\) tokens, each annotated with noisy labels across 28 \glspl{ave} plus a background class denoting silence. As baseline, we adopt the simplest \gls{av}-\gls{mim} from~\cite{Tian2018AVE} without Level-II embeddings, using VGG-19 and VGGish for unimodal features (Level I), bidirectional \glspl{rnn} for temporal modeling (Level III), and late fusion via concatenation (Level IV) followed by fully connected layers for token-based category prediction, which achieves 67.20\% average test accuracy when both modalities' streams are fully available. However, we stress that our \texttt{wrapper} is also compatible with other model choices, \emph{e.g.},~\cite{Mahmud2023AVECLIP,Zhou2025Towards}. We adopt: \((L_a,L_v)\!=\!(5120,B_v)\) bits w/ \((D_a,D_v)\!=\!(20,62.5)\,\mathrm{ms}\) and samps. per pkt. \((S_a^p,S_v^p)\!=\!(320,1)\), given \((R_a,R_v)\!=\!(16\,\mathrm{k},16)\,\mathrm{Hz}\), \(B_a\!=\!16\) bits/samp. (PCM) and \(B_v\!=\!8\times224\times224\times3=1204224\) bits/samp. (PNG). Each $O_i$ has \((N_a,N_v)\!=\!(500,160)\) pkts. w/ \((S_a,S_v)\! =\! (16000,16)\) samps. and \((N_a^k,N_v^k) \!=\! (50,16)\) pkts. per token; dims. are \(d_a\!=\!128\), \(d_v\!=\!512\times7\times7\), \(d_a'\!=\!d_v'\!=\!512\), \(d\!=\!1024\). 

\gls{sota} \texttt{wrappers}~\cite{Li2021SpeculativeInference,Wang2023PATCH,Wu2024AdaFlow,Xu2024MLLMInference} wrap the \gls{av}-\gls{mim} to wait for the reference modality, achieving the minimal accuracy--latency point, which all methods share. 
 
\begin{figure}
    \centering
    \input{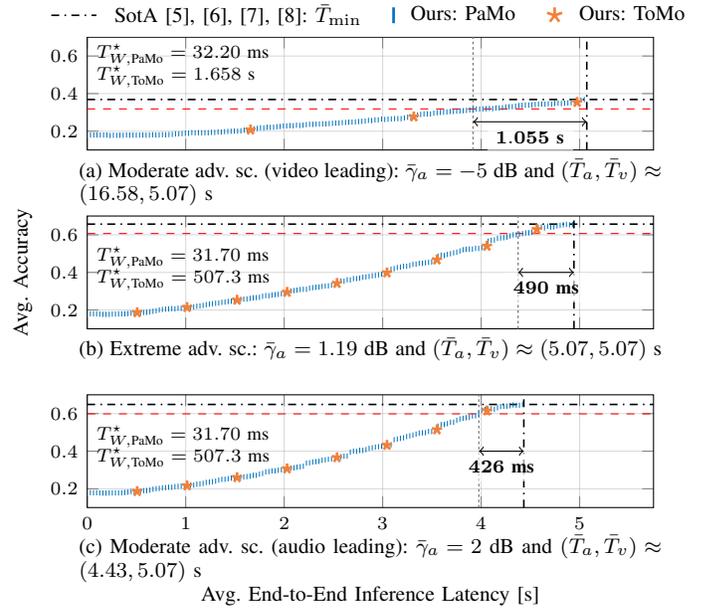}
    \vspace{-1em}
    \caption{
        Average accuracy over the test set as a function of average end-to-end inference latency. Our neuro-inspired, non-blocking inference \texttt{wrapper} is shown by its two design variants, PaMo and ToMo, which correspond to different strategies for defining the \gls{twi}. We compare them to the average minimum end-to-end latency \(\bar{T}_{\rm min} = \mathbb{E}[T_{i,{\rm min}}]\) of \gls{sota} methods~\cite{Li2021SpeculativeInference,Wang2023PATCH,Wu2024AdaFlow,Xu2024MLLMInference}, across varying auditory \gls{snr} values \(\bar{\gamma}_a\) and fixed visual \gls{snr} \(\bar{\gamma}_v = 0\,\mathrm{dB}\). We denote \(\bar{T}_s = \mathbb{E}[T_{i,s}]\) for \(s \in \{a,v\}\) as the average total transmission time per modality. {Our methods enable explicit accuracy--latency trade-offs, which \gls{sota} approaches cannot provide.} The `\,\textcolor{red}{\hdashrule[0.5ex][c]{4.5mm}{0.5pt}{1mm 0.5mm}}' lines indicate a 5\%-drop-margin of accuracy w.r.t. SotA. 
    }
    \label{fig:results}
\end{figure}

We simulate challenging communication conditions by setting outage probabilities \(\varepsilon_a \!=\! \varepsilon_v \!=\! 50\%\). Further, we set the audio-collecting source to be simpler and more resource-constrained than the video one by adopting 5G NR communication parameters representative of \gls{emtc} and \gls{embb} traffic~\cite{3gpp_tr_36101_38101}. We set \(W_v \!=\! 100\,\mathrm{MHz}\) for visual transmission over nominal \gls{embb} conditions and \(W_a \!=\! 1.08\,\mathrm{MHz}\) for auditory transmission using \gls{emtc}. Then, we fix \(\bar{\gamma}_v \!=\! 0\,\mathrm{dB}\) and vary \(\bar{\gamma}_a\). By adjusting \(\bar{\gamma}_a\), we evaluate three possible scenarios for the arrival time of the modalities at the \gls{ma}: in Fig.~\ref{fig:results}(a), video is the reference modality; in Fig.~\ref{fig:results}(b), audio and video have comparable communication delays and either can be the reference; and in Fig.~\ref{fig:results}(c), audio is the reference.

Figure~\ref{fig:results} presents a comparative analysis between \gls{sota} \texttt{wrapper} and our proposed neuro-inspired approach. Given that auditory information is generally more discriminative than visual cues in the \gls{avel} dataset~\cite{Tian2018AVE}, the \gls{sota} methods achieve their highest accuracy relative to the amount of data available of 64.87\% in Fig.~\ref{fig:results}(c). Correspondingly, as the auditory packet rate increases with higher values of \(\bar{\gamma}_a\), the accuracy of our PaMo and ToMo variants exhibits a slightly steeper accuracy improvement from Figs.~\ref{fig:results}(a) and~\ref{fig:results}(b) to~\ref{fig:results}(c), where each represents different strategies for selecting the \gls{twi}. To quantify the accuracy--latency tradeoff, we use the \gls{sota} accuracy as a baseline and define a tolerable accuracy loss of 5\% absolute, indicated by the dashed red horizontal line. Within this threshold, our method can reduce latency by \((1055, 490, 426)\,\mathrm{ms}\) across the three scenarios, with the greatest gain when audio lags, reflecting its higher discriminative power. The main difference between PaMo and ToMo variants lies in their computational and inference profiles: ToMo triggers the \gls{av}-\gls{mim} pipeline less frequently, while PaMo provides finer-grained accuracy--latency control. Overall, our neuro-inspired paradigm dynamically adapts to communication delay uncertainties and effectively explores the accuracy-latency tradeoff curve unlike reference-modality-based \gls{sota} methods.

 
%
%


\section{Conclusions and Future Work}
\label{sec:conclusions}
We propose a novel neuro-inspired, non-blocking inference paradigm that primarily leverages adaptive \glspl{twi} to maintain computational temporal coherence for real-time inference in distributed multimodal systems. By modeling communication uncertainties and statistically optimizing \glspl{twi} accordingly, our approach achieves finer control over the accuracy--latency tradeoff; an ability absent in current \gls{sota} methods. Future work may extend this framework by exploring alternative optimization techniques, incorporating additional delay sources, and integrating synchronization, speculation, and rollback mechanisms from \gls{sota} while adapting them to operate over \glspl{twi} and relaxing our current assumptions of perfect synchronization and zero-data imputation.

\bibliographystyle{IEEEtran}

\end{document}